\documentclass[letterpaper, 10 pt, conference]{ieeeconf}  

\IEEEoverridecommandlockouts                              
\usepackage{amsmath,amssymb,amsfonts, mathtools}
\usepackage{graphicx}
\usepackage{bm}
\usepackage{epstopdf}
\usepackage{multicol}
\usepackage{pgfplots}
\usepackage{multirow}
\usepackage{tensor}
\usepackage{caption}
\usepackage{overpic}
\usepackage{rotating}
\usepackage{verbatim}
\usepackage{algorithm,algorithmic}
\usepackage{color}
\usepackage{soul}
\usepackage{xspace}
\usepackage{todonotes}
\usepackage{url}
\usepackage{float}
\usepackage{helvet}
\usepackage{courier}
\usepackage{type1cm}
\usepackage{makeidx}
\usepackage[bottom]{footmisc}
\usepackage{tikz}
\usepackage{xcolor}
\usetikzlibrary{shapes,arrows}
\let\labelindent\relax
\usepackage[inline]{enumitem}
\usepackage[caption=false, font=footnotesize]{subfig}

\pgfplotsset{compat=newest}
\DeclareMathOperator*{\argmax}{arg\!\max}
\DeclareMathOperator*{\argmin}{arg\!\min}
\makeatletter
\renewcommand*\env@matrix[1][*\c@MaxMatrixCols c]{%
	\hskip -\arraycolsep
	\let\@ifnextchar\new@ifnextchar
	\array{#1}}
\makeatother

\title{\LARGE \bf
	Learning Generalizable Robot Skills from Demonstrations \\ in Cluttered Environments 
}

\author{M. Asif Rana, Mustafa Mukadam, S. Reza Ahmadzadeh, Sonia Chernova, and Byron Boots
	\thanks{
	All authors are affiliated with the Institute for Robotics and Intelligent Machines (IRIM), Georgia Institute of Technology, Atlanta, GA. {Email: \tt\small \{asif.rana, mmukadam3, reza.ahmadzadeh, chernova, bboots\}@gatech.edu.}}
}

\begin{document}

\maketitle
\thispagestyle{empty}
\pagestyle{empty}

\begin{abstract}
Learning from Demonstration (LfD) is a popular approach to endowing robots with skills without having to program them by hand. Typically, LfD  relies on human demonstrations in clutter-free environments. This prevents the demonstrations from being affected by irrelevant objects, whose influence can obfuscate the true intention of the human or the constraints of the desired skill. However, it is unrealistic to assume that the robot's environment can always be restructured to remove clutter when capturing human demonstrations. To contend with this problem, we develop an \emph{importance weighted} batch and incremental skill learning approach, building on a recent inference-based technique for skill representation and reproduction. Our approach reduces unwanted environmental influences on the learned skill, while still capturing the salient human behavior. We provide both batch and incremental versions of our approach and validate our algorithms on a 7-DOF JACO2 manipulator with \emph{reaching} and \emph{placing} skills.
\end{abstract}

\section{Introduction}\label{sec:intro} 
Intelligent and cooperative robots must be capable of adapting to novel tasks in dynamic, unstructured environments. This is a challenging problem to address; it requires a robot to possess a diverse set of skills that may be difficult to hand-specify or pre-program. 
Learning from demonstration (LfD) has proven an effective tool in approaching such problems~\cite{argall2009survey}. To acquire a desired skill, LfD approaches generally involve learning a skill model  from a set of demonstrations provided by a human. 
The model can then be queried to reproduce the skill in novel reproduction environments with additional skill constraints. Common examples of constraints include new start/goal states, or new obstacle configurations that constrain the set of possible trajectories. 
 LfD techniques generally differ in the manner in which the skill is represented, learned, and reproduced.


Most prior LfD approaches~\cite{ijspeert2013dynamical,calinon2007learning,khansari2011learning,paraschos2013probabilistic} are based on the assumption that demonstrations can be performed in uncluttered, minimally constrained environments.  The presence of clutter in the demonstration environments can introduce additional constraints on human demonstrations that are unrelated to the target skill or the underlying human intent. If unaccounted for, this can lead to suboptimal skill models. However, restructuring the world to remove clutter is often impractical, which limits the viability of such approaches.

\begin{figure}
	\centering
	\includegraphics[trim={0cm 1cm 0cm 0cm}, clip, width=0.35\textwidth]{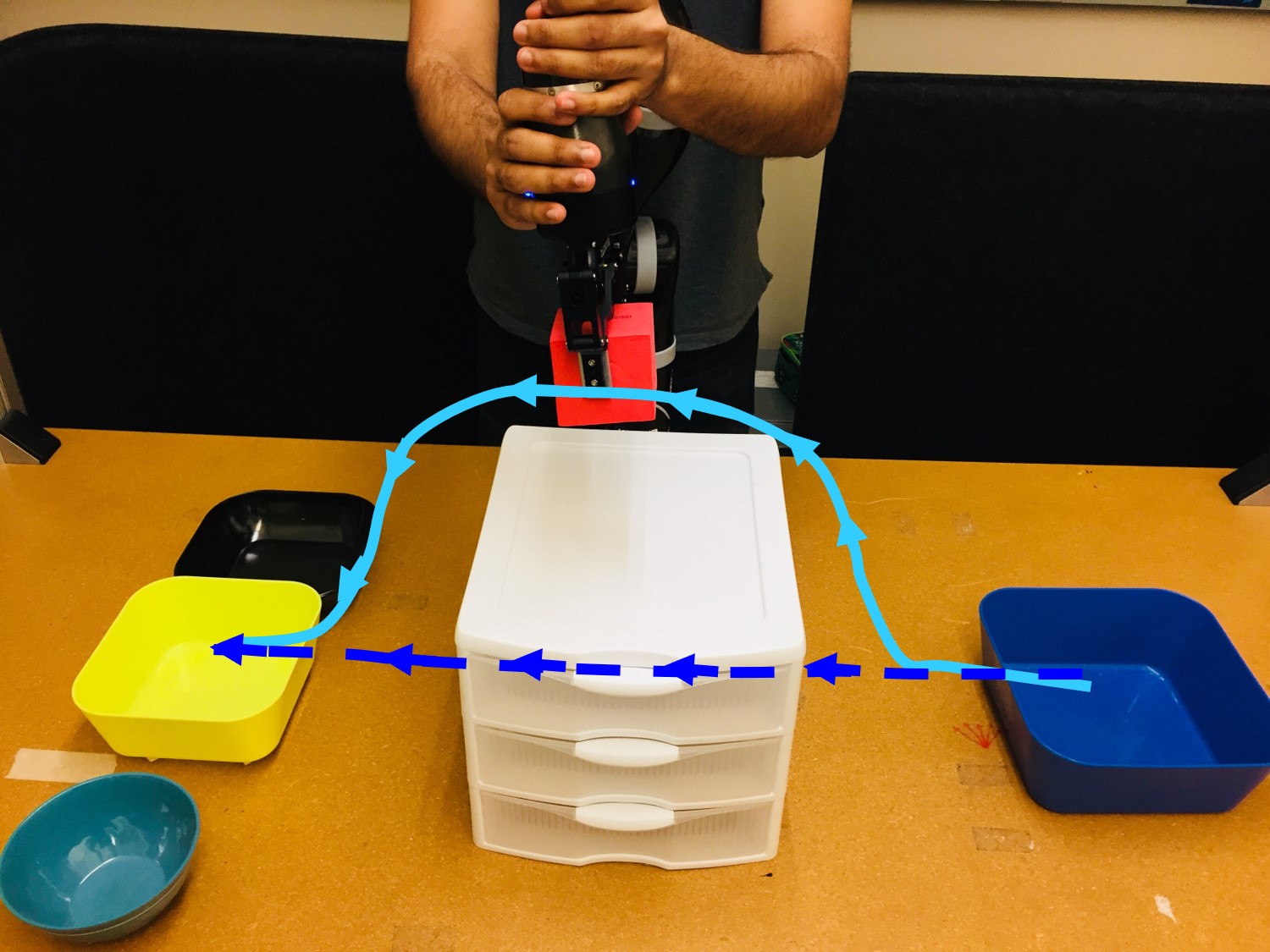}
 	\caption{\small
	A human is demonstrating a placing skill, which involves placing the (red) cube from the (blue) bowl on the right in to one of the three bowls on the left. The figure contrasts the demonstrated trajectory (light blue), which is influenced by an obstacle (drawer) in the environment, with the intended straight-line trajectory (dark blue) in the absence of the obstacle.}
	\vspace{-0.60cm}
	\label{fig:cover}
\end{figure}

In this work, we tackle the problem of learning skills from a set of demonstrations, which can be partially or fully influenced by the presence of obstacles (see Fig.~\ref{fig:cover}). 




To contend with obstacles during training, we present \emph{importance weighted skill learning}. Specifically, we adopt and extend the inference-based view of skill reproduction as proposed by Rana et al.~\cite{rana2017towards} with Combined Learning from Demonstration And Motion Planning (CLAMP). CLAMP provides a principled approach for generalizing robot skills to novel situations, including avoiding unknown obstacles present in the reproduction environment. When reproducing a desired skill, trajectories are generated to be \emph{optimal} with respect to the demonstrations while remaining \emph{feasible} in the given reproduction environment.

\begin{figure}
	\centering
	\includegraphics[trim={0cm 4cm 0cm 4cm}, clip, width=0.44\textwidth]{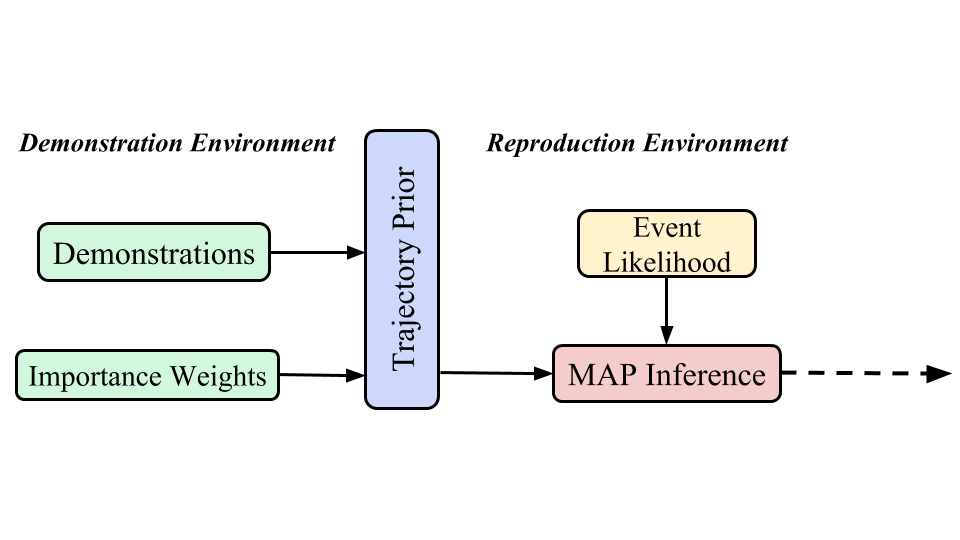}
	\caption{\small{An overview of our approach. In the demonstration environment, the human demonstrations and the associated importance weights are collected. The trajectory prior acts as the skill model. Conditioning this prior on the likelihood of events specified by the reproduction scenario gives the posterior.}}
	\vspace{-0.6cm}
	\label{fig:blk_diag}
\end{figure}

We extend CLAMP to utilize demonstrations from cluttered environments through importance weighted skill learning (see Fig.~\ref{fig:blk_diag}), which rates the importance of demonstration trajectories while learning the skill model. We propose an importance weighting function that assigns lower importance to parts of demonstrations that are more likely to be influenced by obstacles.
We present batch and incremental versions of our algorithm: batch learning is useful when the set of initial demonstrations are sufficient for learning a reasonable skill model, while incremental learning is useful in scenarios that require refinement of the skill model as new demonstrations in new environments become available.

We validate our approach on 
a 7-DOF JACO2 manipulator with \emph{reaching} and \emph{placing} skills. In all the experiments, we evaluate the approach by providing demonstrations in cluttered environments and then changing the environments for reproduction.

\section{Related Work}
Many existing approaches to trajectory-based LfD address the problem of avoiding obstacles in the reproduction scenario. Some approaches add obstacle avoidance in the skill reproduction phase as a reactive strategy~\cite{pastor2009learning, park2008movement, khansari2012dynamical}, while others carry out motion planning or trajectory optimization~\cite{ye2011demonstration,osa2017guiding,koert2016demonstration,rana2017towards}. In all these approaches, the skill model is learned from demonstrations that are not affected by obstacles. Any constraints or costs associated with obstacles are typically present during reproduction only. However, in an obstacle-rich environment, the demonstrations themselves are likely to be influenced by the presence of obstacles, which could have repercussions during skill reproduction.

There have been a few attempts to address the problem of learning skills from demonstrations in cluttered environments. For example, \cite{rai2014learning, gams2015learning} learn a dynamic movement primitive (DMP) as well as a coupling term for obstacle avoidance from demonstrations. These approaches suffer from two major problems. First, since DMPs follow a single demonstration, they fail to learn potentially different ways of executing the skill, thereby limiting its robustness in new scenarios. Second, due to the reactive nature of the obstacle avoidance strategy, the reproduced trajectory does not necessarily preserve the shape of the motion in the presence of obstacles.  Ghalamzan et al.~\cite{ghalamzan2015incremental}, proposed an approach based on learning a cost functional from human demonstrations. This cost functional is dependent on two components: the deviation from the mean of the demonstrations, and the distance from obstacles in the environment. Parameters of both these components are estimated from human demonstrations. A major drawback of this approach is the assumption that the mean of the demonstrations sufficiently expresses the demonstrated skill. This assumption however stands invalid for skills which can be executed in multiple ways and hence requires a more expressive skill model.





Our proposed method is based on learning an underlying stochastic dynamical system from demonstrations. 
Depending on the part of the state-space the robot lies in, this dynamical system is able to generate different ways of executing a learned skill. 
We make use of importance weighting to discount the effect of obstacles that are present when the demonstrations are provided. Specifically, the parts of demonstrations in the vicinity of obstacles are penalized to account for their deviation from the desired skill or the human intention. 

\section{Combined Learning from Demonstration and Motion Planning}
We adopt the probabilistic inference view on learning from demonstration which has been previously employed in CLAMP~\cite{rana2017towards}. 

\subsection{Skill Reproduction as Probabilistic Inference}
Skill reproduction using CLAMP is performed by \emph{maximum a posteriori} (MAP) inference given a trajectory prior and event likelihoods in the reproduction environment.

\subsubsection*{\textbf{Trajectory Prior}}\label{sec:prior}
The trajectory prior or the skill model represents a distribution over robot trajectories. A trajectory is defined as a finite collection of $D$-dimensional robot states $\bm{x}_i \in \mathbb{R}^D$ at time $t_i$, $0 \leq i \leq N$. The prior is given by a joint Gaussian distribution over the robot states,
\begin{align}\label{eq:original_prior}
p(\bm{x}) \propto \exp \{ - \frac{1}{2} \| \bm{x} - \bm{\mu} \|^{2}_{\bm{\mathcal{K}}} \},
\end{align}
where,
\begin{equation*}
\bm{x} \doteq [\bm{x}_0, \bm{x}_1, \dots, \bm{x}_N]^T
\vspace{-1mm}
\end{equation*}
\begin{equation*}
\boldsymbol{\mu} \doteq [\bm{\mu}(t_0), \bm{\mu}(t_1), \dots, \bm{\mu}(t_N)]^T,\
\boldsymbol{\mathcal{K}} \doteq [\boldsymbol{\mathcal{K}}(t_i,t_j)]\big|_{i j, 0\leq i,j\leq N}.
\end{equation*}
The prior enforces \textit{optimality} by penalizing the optimal trajectory on deviating from the mean of the prior during inference. The trajectory prior is learned from demonstrations.

\subsubsection*{\textbf{Event Likelihood}}
The likelihood function encodes the constraints in the skill reproduction scenario. The constraints are represented as random events $\mathbf{e}$ that the optimal trajectory should satisfy thus enforcing \textit{feasibility} during inference i.e. reproduction. These events, for example, may include obstacle avoidance, or a new start/goal state or via-point. The likelihood function is defined as a distribution in the exponential family,
\begin{align}\label{eq:likelihood}
p( \mathbf{e} | \bm{x}) \propto  \exp \{ - \frac{1}{2} \| \bm{h}(\bm{x};\mathbf{e}) \|^{2}_{\mathbf{\Sigma}} \},
\end{align}
where $\bm{h}(\bm{x};\mathbf{e})$ is a vector-valued cost function with covariance matrix $\bm\Sigma$. The reader is referred to~\cite{mukadam2017continuous, rana2017towards} for more details on these likelihood functions.

\subsubsection*{\textbf{MAP Inference}}
The desired optimal and feasible trajectory that reproduces the skill is then given by,
\begin{align}\label{eq:MAP}
\bm{x}^* &= \argmax_{\bm{x}} \big\{ p( \bm{x} |  \mathbf{e}) \big\} = \argmax_{\bm{x}} \big\{ p(\bm{x})  p( \mathbf{e} | \bm{x})\big\}.
\end{align}

\subsection{Trajectory Prior Formulation}\label{sec:sde}
It is assumed that in CLAMP that robot trajectories for a desired skill are governed by an underlying linear stochastic skill dynamics,
\begin{align}\label{eq:dynamics}\small
\hspace{-2mm} \bm{x}_{i+1} = \bm{\Phi}_{i+1}\bm{x}_{i} + \mathbf{u}_{i+1} + \mathbf{w}_{i+1},\quad \mathbf{w}_{i+1} \sim \mathcal{N}(\bm{0}, \mathbf{Q}_{i+1}),\hspace{-2mm}
\end{align}
where $\bm{\Phi}_i$ and $\bm{u}_i$ are a time-varying transition matrix and a bias term, respectively, and $\mathbf{w}_i$ is additive white noise with time-varying covariance $\mathbf{Q}_i$. The trajectory prior can be generated by taking the first and second order moments of the solution to this dynamics. This Markovian dynamics yields an exactly sparse precision matrix (inverse covariance)~\cite{Mukadam-ICRA-16, barfoot2014batch} inducing structure in the trajectory prior in~\eqref{eq:original_prior}, which enables efficient learning and inference. The problem of learning the trajectory prior is equivalent to estimating the underlying stochastic dynamics. 

While learning the trajectory prior, CLAMP assumes all available demonstrations are free from external influences, and therefore captures the true human intent or skill constraints. However, in the presence of such influences, this assumption no longer holds and the learned prior is suboptimal.

\section{Importance weighted skill learning}\label{sec:skill_learning}
In this section, we introduce importance weighting when learning the prior to 
exclude the effects of unwanted influences during demonstrations. We seek to estimate the parameters of the skill dynamics model in~\eqref{eq:dynamics} from demonstrations. As a preliminary step, lets re-write~\eqref{eq:dynamics} as follows,
\begin{equation}\label{eq:dynamics_new}
	\bm{x}_{i+1} = \bm{\tilde{\Phi}}_{i+1}\bm{\tilde{x}}_i + \mathbf{w}_{i+1},\ \ \ \ \ \mathbf{w}_{i+1} \sim \mathcal{N}(\bm{0}, \mathbf{Q}_{i+1})
\end{equation}
where,
\begin{equation*}
\bm{\tilde{x}}_{i} = \left[
    \begin{array}{c}
       \mathbf{1} \\
		 \bm{x}_i
     \end{array} 
\right], \ \ \ \ \  
\bm{\tilde{\Phi}}_{i+1} = 
\begin{bmatrix}[cc]
  	\mathbf{u}_{i+1} & \bm{\Phi}_{i+1}
\end{bmatrix}
\end{equation*}
We additionally define an importance weighting function as $w: \mathbb{R}^d \mapsto \mathbb{R}$. The importance weighting function should give higher weights to robot states that are less likely to deviate from the skill constraints or the true human intent. While this importance weighting formulation can be used in other contexts too, in this paper we define a specific form of importance weighting to account for the influence of unwanted obstacles in the demonstration environment. The exact form of this environment-dependent obstacle weighting function is presented in Section \ref{sec:weight_function}.

\subsection{Batch Skill Learning}\label{sec:batch_learning}
Let's assume the availability of $K$ trajectory demonstrations, with the $k^{\textrm{th}}$ demonstration defined as $\bm{x}^k = [\bm{x}_0^k,\bm{x}_1^k,\dots,\bm{x}_N^k]^T$. For each discrete time interval $(t_i, t_{i+1}]$, the inputs are collected into a matrix $\bm{\tilde{X}}_i = [\bm{\tilde{x}}_i^1, \bm{\tilde{x}}_i^2, \dots, \bm{\tilde{x}}_i^K]$ while the corresponding targets into a matrix $\bm{X}_{i+1} = [\bm{x}_{i+1}^1, \bm{x}_{i+1}^2, \dots, \bm{x}_{i+1}^K]$. Furthermore, the matrix $\mathbf{W}_i = \text{diag}\big(w(\bm{x}_i^1), w(\bm{x}_i^2), \dots, w(\bm{x}_i^K)\big)$ defines a state-dependent importance weight matrix.

The batch skill learning formulation seeks to find $\bm{\tilde{\Phi}}_{i+1}$ and $\mathbf{Q}_{i+1}$, which minimize a regularized squared norm over the provided demonstrations.
\begin{align}\label{eq:loss}
\bm{\tilde{\Phi}}_{i+1}^*, & \mathbf{Q}_{i+1}^* \\
\nonumber &= \argmin_{\bm{\tilde{\Phi}}_{i+1}, \mathbf{Q}_{i+1}} \bigg\{\mathcal{L}(\bm{\tilde{\Phi}}_{i+1}, \mathbf{Q}_{i+1}) \bigg\}\\
\nonumber &= \argmin_{\bm{\tilde{\Phi}}_{i+1}, \mathbf{Q}_{i+1}} \bigg\{\text{tr}\big(\mathbf{Q}_{i+1}^{-1}\mathbf{E}_{i+1}\mathbf{W}_i\mathbf{E}_{i+1}^T\big) + \lambda \|\bm{\tilde{\Phi}}_{i+1}\|_F^2\bigg\}
\end{align}
where $\mathbf{E}_{i+1} = \bm{X}_{i+1} - \bm{\tilde{\Phi}}_{i+1}\bm{\tilde{X}}_{i}$ defines the error matrix, and $\lambda$ is a regularization coefficient.

The solution to the batch skill learning problem in \eqref{eq:loss} is given by the weighted ridge regression estimate,
\begin{align}
\bm{\tilde{\Phi}}_{i+1}^* &= 
\bm{X}_{i+1}^T\mathbf{W}_i\bm{\tilde{X}}_i\big(\bm{\tilde{X}}_i \mathbf{W}_i\bm{\tilde{X}}_i^T + \lambda\mathbf{I}\big)^{-1},\\
\mathbf{Q}_{i+1}^* &= \frac{1}{z}\mathbf{E}_{i+1}^*\mathbf{W}_i{\mathbf{E}_{i+1}^*}^{T},\\
\nonumber z &= \frac{\text{tr}(\mathbf{W}_i)^2 - \text{tr}(\mathbf{W}_i^T\mathbf{W}_i)}{\text{tr}(\mathbf{W}_i)}.
\end{align}

\subsection{Incremental Skill Learning}\label{sec:incremental_learning}
The batch skill learning procedure assumes that there are enough demonstrations available to learn an optimal skill model. However, as more demonstrations are aggregated over time, possibly in different  environments, it is desirable to refine the model since more data provides a better estimate of the skill. To achieve this, we propose incremental weighted skill learning. 


Our incremental skill learning procedure is based on Bayesian inference. In this formulation, we maintain a joint probability distribution over the unknown skill dynamics parameters. Every time a new demonstration is collected, a posterior over the skill dynamics parameters is calculated 
\begin{align}
\begin{multlined}
p(\bm{\tilde{\Phi}}_{i+1},\mathbf{Q}_{i+1} | \mathcal{D}^{1:k})\\ = p(\mathcal{D}^{k}| \bm{\tilde{\Phi}}_{i+1},\mathbf{Q}_{i+1})p(\bm{\tilde{\Phi}}_{i+1},\mathbf{Q}_{i+1} | \mathcal{D}^{1:k-1}),
\end{multlined}
\end{align}
where $\mathcal{D}^{1:k} = \{\{\bm{\tilde{x}}^1_i, \bm{{x}}^1_{i+1}\}, \{\bm{\tilde{x}}^2_i, \bm{{x}}^2_{i+1}\}, \dots, \{\bm{\tilde{x}}^k_i, \bm{{x}}^k_{i+1}\}\}$. At any stage, the mode of the posterior distribution provides an estimate of the unknown parameters.


\subsubsection*{\textbf{Skill Dynamics Distribution}}
The joint probability distribution over the unknown parameters $\bm{\tilde{\Phi}}_{i+1}$ and $\mathbf{Q}_{i+1}$ is given by
\begin{align}\label{eq:prior_dynamics}
p(\bm{\tilde{\Phi}}_{i+1},\mathbf{Q}_{i+1}) = p(\bm{\tilde{\Phi}}_{i+1}|\mathbf{Q}_{i+1})p(\mathbf{Q}_{i+1}),
\end{align}
where,
\begin{align}
p(\bm{\tilde{\Phi}}_{i+1}|\mathbf{Q}_{i+1}) &= \mathcal{MN}(\bm{M}_{i+1}, \mathbf{Q}_{i+1}, \bm{R}_{i+1}),\\
p(\mathbf{Q}_{i+1}) &= \mathcal{W}^{-1}(\bm{V}_{i+1}, \nu_{i+1}),
\end{align}
$\mathcal{MN}$ refers to a matrix-normal distribution with matrix-valued mean $\bm{M}_{i+1}$ and covariances $\mathbf{Q}_{i+1}$ and $\bm{R}_{i+1}$ for the rows and columns respectively. $\mathcal{W}^{-1}$ refers to an inverse-Wishart distribution with positive definite scale matrix $\bm{V}_{i+1}$ and $\nu_{i+1}$ degrees of freedom. Note that matrix-normal and inverse-Wishart distributions are generalizations of the normal and inverse-gamma distributions respectively to the multivariate case.


\subsubsection*{\textbf{Demonstration Likelihood}}
The likelihood of observing the input-target pair from the $k^{th}$ demonstration under the stochastic dynamics~\eqref{eq:dynamics_new} is given by
\begin{align}\label{eq:likelihood_dynamics}
p(\mathcal{D}^{k}| \bm{\tilde{\Phi}}_{i+1},\mathbf{Q}_{i+1}) &\doteq 
p(\bm{x}^{k}_{i+1}|\bm{\tilde{x}}^k_i, \bm{\tilde{\Phi}}_{i+1}, \mathbf{Q}_{i+1}) \\
\nonumber &\propto \exp\bigg\{-\frac{1}{2}(w_i^k\mathbf{Q}_{i+1}^{-1}\mathbf{e}_{i+1}^k{\mathbf{e}_{i+1}^k}^T)\bigg\}.
\end{align} 
where $\mathbf{e}_{i+1}^k = \bm{x}^k_{i+1} - \bm{\tilde{\Phi}}_{i+1}\bm{\tilde{x}}_i^k$ and $w_i^k = w(\bm{x}^k_i)$. Note that the likelihood is scaled by the weight in order to incorporate the importance weighting.

\subsubsection*{\textbf{Skill Dynamics Inference}}
The skill dynamics parameters after assimilation of $k$ demonstrations is given by the mode of the joint posterior distribution (\textit{maximum a posteriori}),
\begin{align}\label{eq:map_dynamics}
& \bm{\tilde{\Phi}}_{i+1}^k, \mathbf{Q}_{i+1}^k
= \argmax_{\bm{\tilde{\Phi}}_{i+1}, \mathbf{Q}_{i+1}} \bigg\{ p(\bm{\tilde{\Phi}}_{i+1},\mathbf{Q}_{i+1}|\mathcal{D}^{1:k})\bigg\}.
\end{align}
Due to the properties of matrix-normal and inverse Wishart distributions, the mode of the joint distribution turns out to be equivalent to the product of the modes of the two conditional distributions~\cite{minka2000bayesian},
\begin{align}
\bm{\tilde{\Phi}}_{i+1}^k &= \argmax_{\bm{\tilde{\Phi}}_{i+1}} \big\{ p(\bm{\tilde{\Phi}}_{i+1}|\mathbf{Q}_{i+1}, \mathcal{D}^{1:k}) \big\}= \bm{M}_{i+1}^k\\
\mathbf{Q}_{i+1}^k &= \argmax_{\mathbf{Q}_{i+1}}\big\{p(\mathbf{Q}_{i+1}|\mathcal{D}^{1:k})\} = \frac{1}{\nu_{i+1}^k + D + 1} \bm{V}_{i+1}^k.
\end{align}
Furthermore, the parameters of the conditional distributions are governed by the following update laws,
\begin{align*}
\bm{R}_{i+1}^k &= w_i\bm{\tilde{x}}_i\bm{\tilde{x}}_i^T + \bm{R}_{i+1}^{k-1}\\
\bm{M}_{i+1}^k &= (w_i\bm{x}_{i+1}\bm{\tilde{x}}_i^T + \bm{M}_{i+1}^{k-1}\bm{R}_{i+1}^{k-1})(\bm{R}_{i+1}^k)^{-1}\\
\bm{V}_{i+1}^k  &=
\begin{multlined}
\bm{V}_{i+1}^{k-1}
+ w_i(\bm{x}_{i+1}-\bm{M}_{i+1}^k \bm{\tilde{x}}_i)(\bm{x}_{i+1}-\bm{M}_{i+1}^k \bm{\tilde{x}}_i)^T \\
+ (\bm{M}_{i+1}^k - \bm{M}_{i+1}^{k-1})\bm{R}_{i+1}^{k-1}(\bm{M}_{i+1}^k - \bm{M}_{i+1}^{k-1})^T 
\end{multlined}\\
\nu_{i+1}^k    &= 1 + \nu_{i+1}^{k-1}
\end{align*}
The incremental learning procedure is initialized with a prior joint distribution $p(\bm{\tilde{\Phi}}_{i+1},\mathbf{Q}_{i+1}|\phi)$. The Gaussian comonent of the joint prior is selected to be the ridge regression prior, that is, $\bm{M}_{i+1}^0 = \mathbf{0}$ and $\bm{R}_{i+1}^0 = \frac{1}{\alpha}\bm{I}$. The inverse Wishart component is selected to be an uninformed prior, with $\bm{V}_{i+1}^0 = \frac{1}{\beta}\bm{I}$ and  $\nu_{i+1}^0 = \frac{1}{\beta}$. Here $\alpha$ and $\beta$ are positive scalars. In our implementation, we set $\alpha=\beta=10^{10}$. Note that smaller values of these scalars makes the prior too strict, which restrains the skill model from fitting the data well.

\section{Environment-dependent importance weighting function}\label{sec:weight_function}
In this section, we define the importance weighting function to enable skill learning from demonstrations, which may be provided in the presence of obstacles in the environment. The weighting function gives lower importance to the parts of a demonstration which are more likely to be influenced by the presence of an obstacle and therefore deviate from the  intent of the human.

\begin{figure}
\includegraphics[trim={0cm 0cm 0cm 0cm},clip, width=0.58\linewidth]{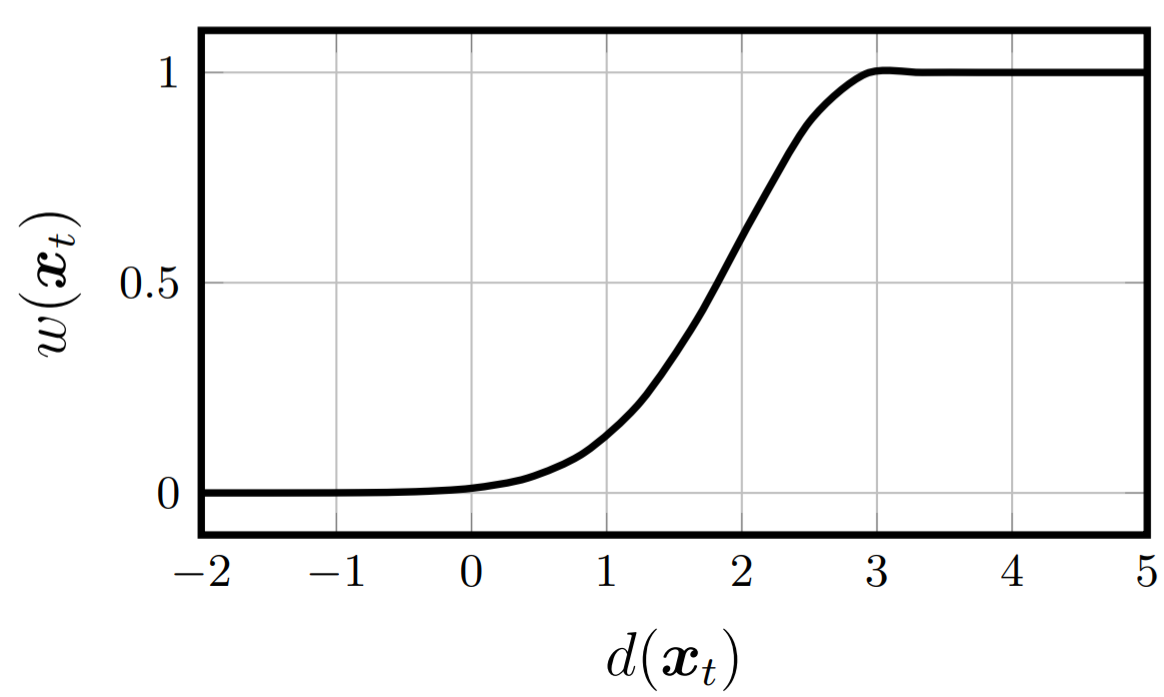}
\hfill
\includegraphics[trim={0cm 0cm 0cm 0cm},clip, width=0.39\linewidth]{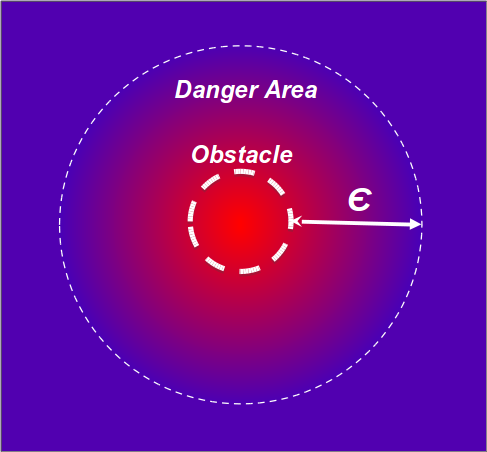}
	\caption{\small{An illustration of an importance weight function parameterized by $\epsilon=3$ and $\sigma_{obs}=1$ (left) and a signed distance field (right). The importance weight levels at 1 outside the danger area, and decays down to zero inside with the slope governed by $\sigma_{obs}$.}}
	\vspace{-0.5cm}
	\label{fig:obs_weight}
\end{figure}

We hypothesize that the parts of demonstrations closer to obstacles are influenced by the obstacles and therefore fail to satisfy the skill constraints. Conversely, partial trajectories farther away from obstacles are more likely to satisfy the skill constraints and  should be given more importance. For a given state $\bm{x}_i$, we define the importance weight to be equivalent to the likelihood of staying collision-free~\cite{mukadam2017continuous}. For this likelihood function, we first define a hinge loss function
\[ c(\bm{x}_i) = \begin{cases} 
-d(\bm{x}_i) + \epsilon & d(\bm{x}_i)\leq \epsilon \\
\ \ 0 & d(\bm{x}_i) > \epsilon
\end{cases},
\]
where $d(\cdot)$ is the signed distance from the closest obstacle in an environment and $\epsilon$ specifies the `danger area' around the obstacle. With this hinge loss, we assume that an obstacle affects a state only when it is within the danger area around the obstacle. Outside of this danger area, the obstacle has no influence on the state. The importance weight itself is given by a function in the exponential family,
\begin{equation}\label{eq:weight_fcn}
w(\bm{x}_i) = \exp\bigg\{ -\frac{c(\bm{x}_i)^2}{2\sigma_{obs}^2}  \bigg\},
\end{equation}
where the parameter $\sigma_{obs}$ dictates the rate of decay of the importance weight for states within the 'danger area'. The smaller the value of $\sigma_{obs}$, the faster the importance weight will decay down to zero (see Fig~\ref{fig:obs_weight}). 



\section{Experiments}
We evaluate the performance of our method on two different skills\footnote{Accompanying video: https://youtu.be/03r8Tblhq7k}: 1) the \textit{reaching} skill, and 2) the \textit{placing} skill. For both  skills, a human provides multiple demonstrations via kinesthetic teaching on a 7-DOF JACO2 manipulator. The end-effector positions are recorded and  the corresponding instantaneous velocities are estimated by fitting a cubic spline to each demonstration and taking its time derivative. Furthermore, the demonstrations are also time-aligned using dynamic time warping (DTW). To setup the trajectory prior in~\eqref{eq:original_prior}, we define the robot states $\bm{x}_i$ as the vector concatenation of instantaneous robot positions and velocities.

\begin{figure}[!t]
    \centering
  \subfloat{%
        \includegraphics[width=0.99\linewidth, height=0.35\linewidth]{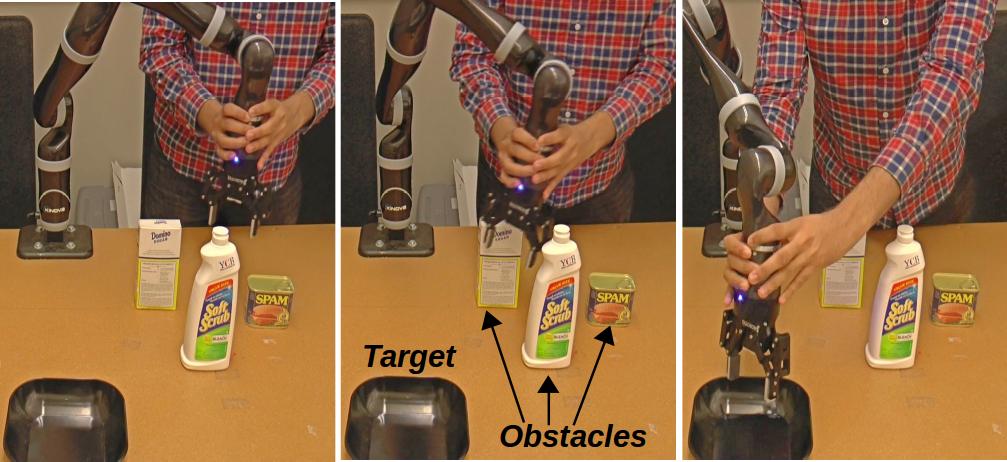}}
        \\
    \subfloat{%
        \includegraphics[width=0.55\linewidth, height=0.40\linewidth]{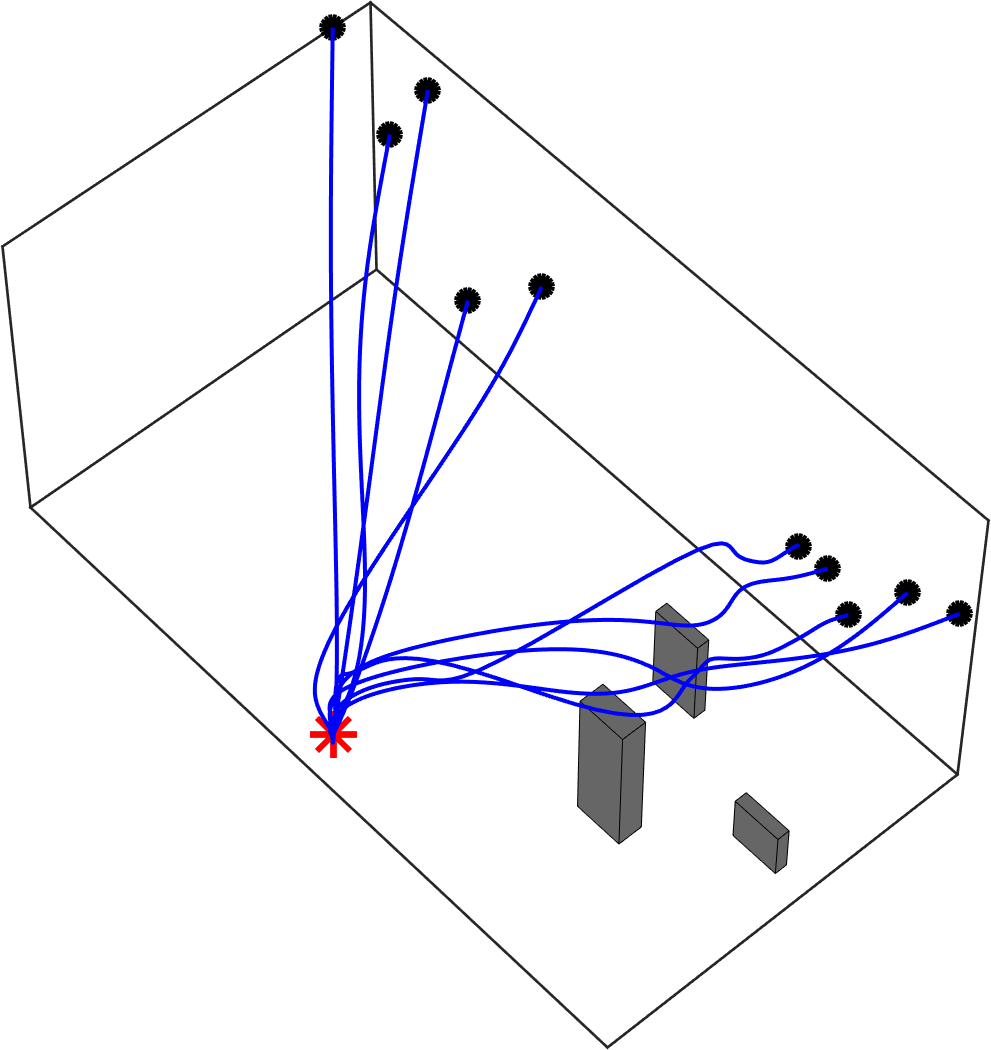}}
        \hfill
  \caption{\small{Human demonstrations for the \emph{reaching} skill. All demonstrations reach the bowl from different initial positions in the presence of three obstacles in the environment. \emph{Top}: Snapshots of a demonstrations avoiding the obstacles. \emph{Bottom}: A 3-D plot showing all the demonstrations and the obstacles.}}
\label{fig:demos_reaching}
\end{figure}

For the \textit{reaching} skill, the goal is to reach an object from different locations. Hence, all the demonstrations share the same goal state while the initial state varies. In the absence of any obstacles in the path, a demonstration follows a nearly straight-line path to the goal. In the presence of obstacles in the path, the demonstrations deviate from this desired path in order to avoid collision with the obstacles. Fig.~\ref{fig:demos_reaching} shows the demonstration environment and the corresponding demonstrations.

\begin{figure}[!t]
    \centering
   \subfloat[without importance weighting]{%
        \includegraphics[width=0.48\linewidth]{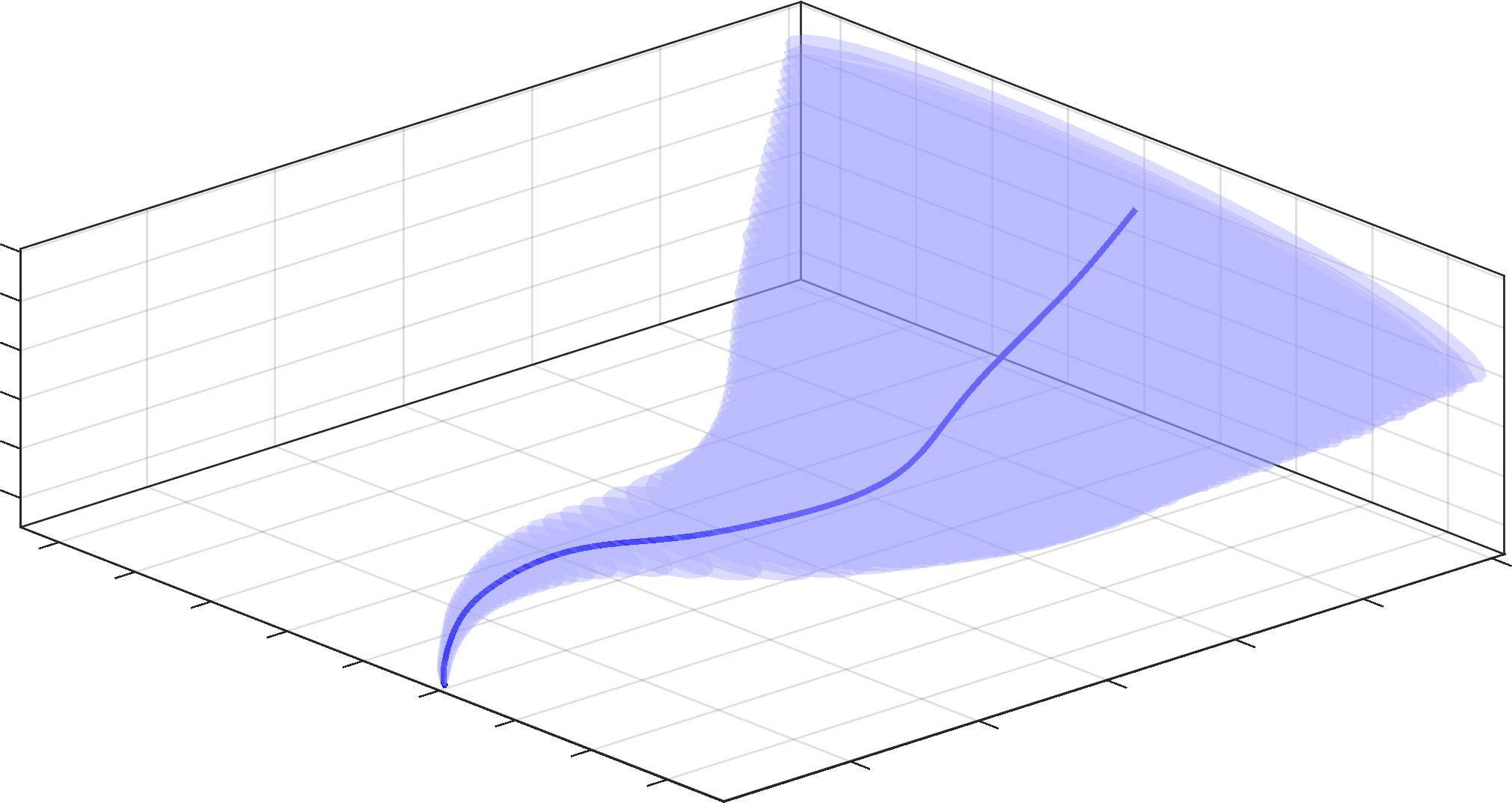}}
    \label{fig:prior_reaching_a}\hfill
  \subfloat[with importance weighting]{%
        \includegraphics[width=0.48\linewidth]{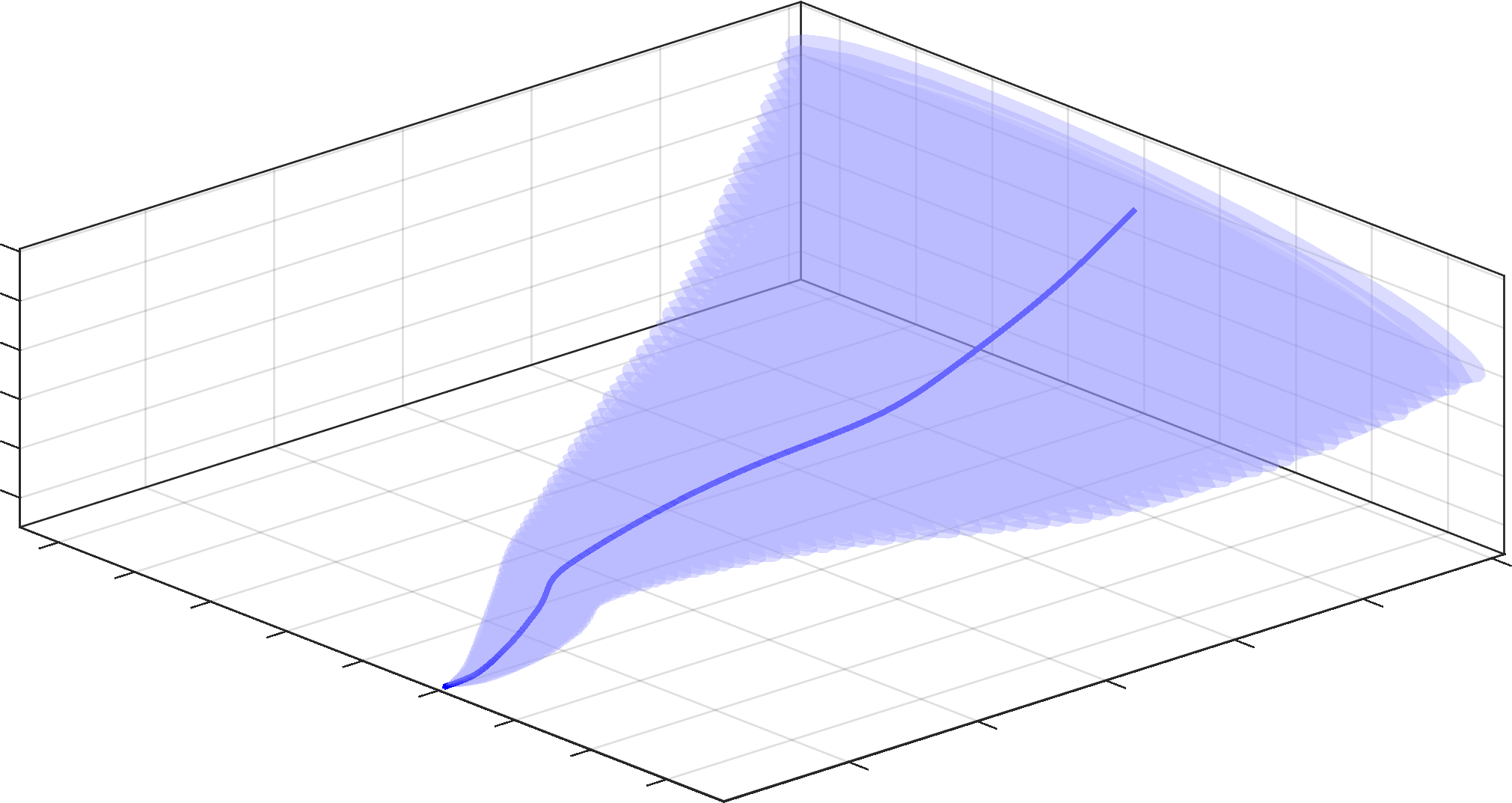}}
     \label{fig:prior_reaching_b} 
  \caption{\small{Trajectory prior visualization for the \emph{reaching} skill. The blue  line is the mean of the prior, and the blue shaded region shows one standard deviation around the mean.}}
  \vspace{-0.5cm}
  \label{fig:prior_reaching} 
\end{figure}

In order to learn the trajectory prior for this skill, we use importance weighted skill learning, as described in Section \ref{sec:batch_learning}. The demonstrations reaching the target from the uncluttered part of the environment represent the true human intent. Therefore, we expect our trajectory prior to be biased towards these demonstrations.  Fig.~\ref{fig:prior_reaching} shows the trajectory distributions (i.e. time-evolving state distributions) encoded in the trajectory priors learned with and without importance weighting. The trajectory distributions are generated by rolling out the stochastic skill dynamics in ~\eqref{eq:dynamics_new} with an initial state distribution given by a Gaussian over the initial demonstration states. The mean of the trajectory distribution generated with importance weighting deviates less from the intended straight-line path, exhibiting the true underlying skill, as compared to the distribution without importance weighting. To enable this, we empirically selected the parameters of the importance weight function in \eqref{eq:weight_fcn}, such that the parts of state-space likely to be under obstacle influence can be successfully downplayed while learning the prior. A value of $\epsilon=0.3m$ and $\sigma_{obs}=0.01m$ provided sufficient bounding region around the obstacles in most cases.

\begin{figure}[!t]
    \centering
  \subfloat{%
        \includegraphics[width=0.64\linewidth]{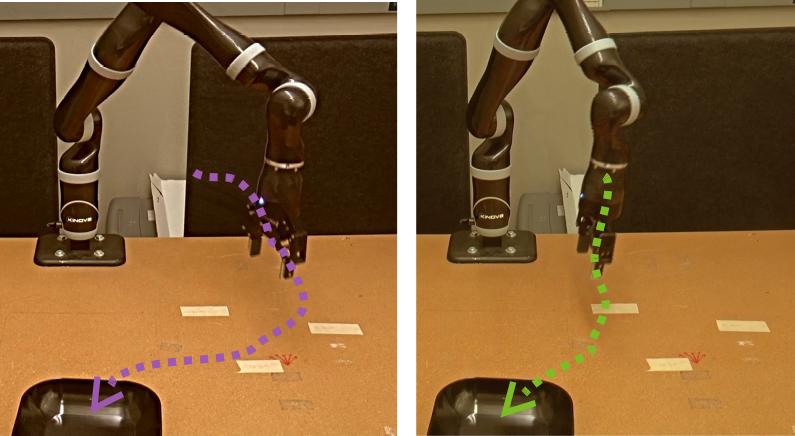}}
        \\
   \subfloat{%
        \includegraphics[width=1.0\linewidth]{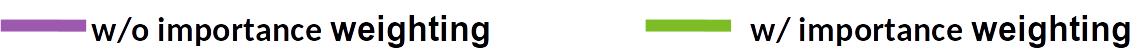}}
        \\
    \vspace{-0.4cm}
   \subfloat{%
        \includegraphics[width=0.32\linewidth,height=0.32\linewidth]{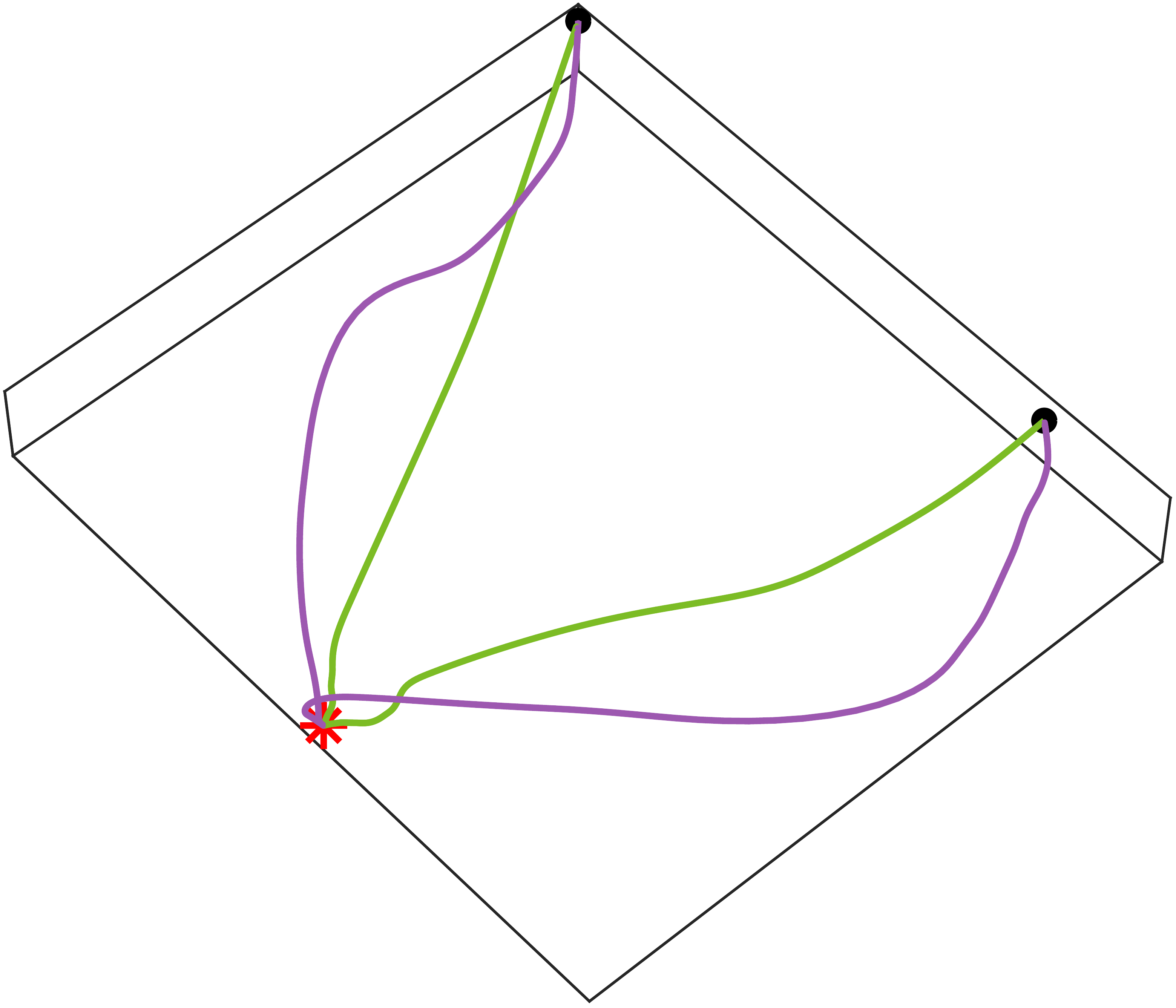}}
    \hfill
  \subfloat{%
        \includegraphics[width=0.32\linewidth]{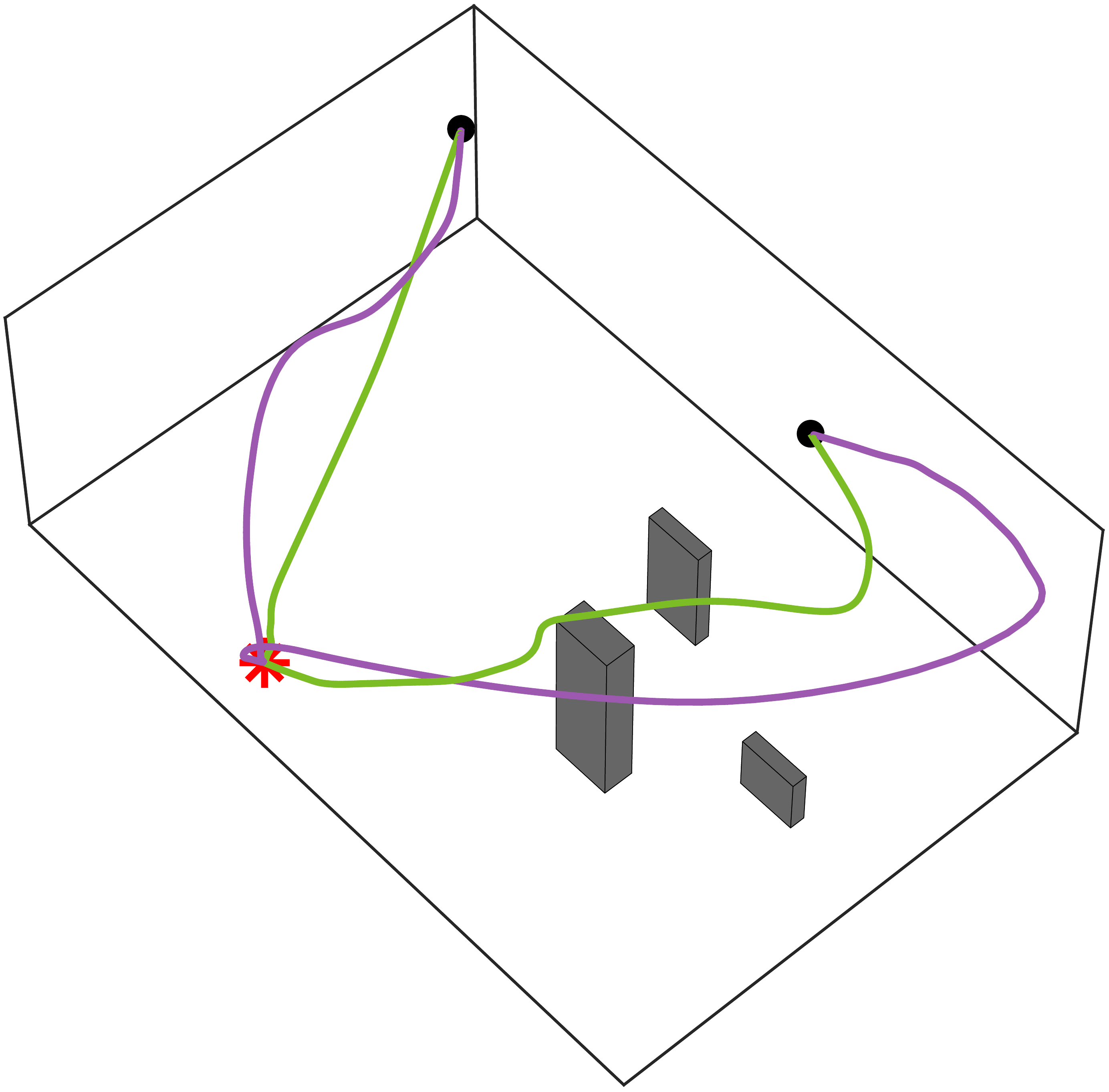}}
    \hfill
  \subfloat{%
        \includegraphics[width=0.32\linewidth]{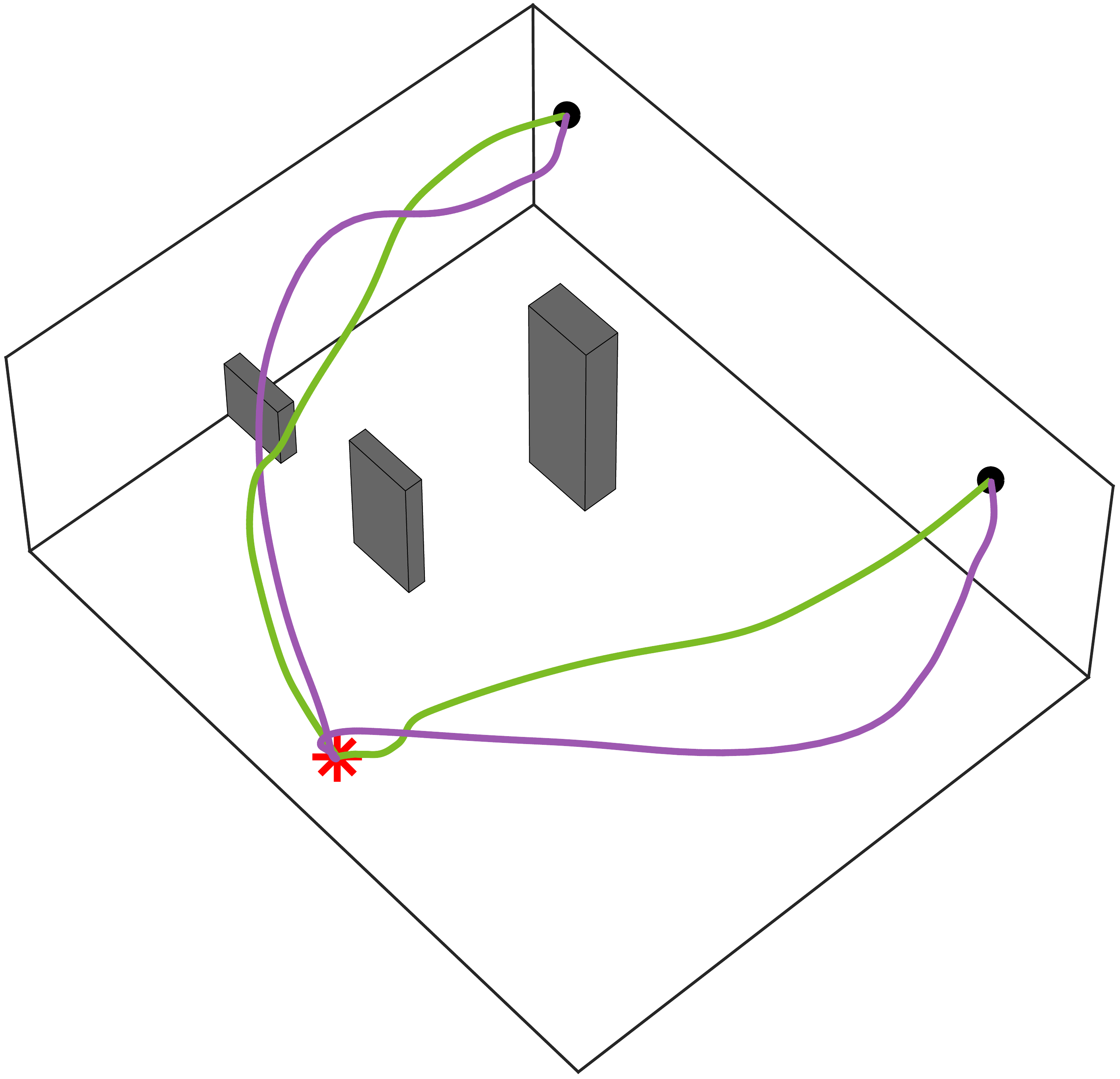}}
  \caption{\small{Trajectories generated by conditioning the priors on two initial positions in three different environments. \emph{Top-left}: Environment without obstacles. \emph{Top-center}: Environment with obstacles at the same locations as demonstrations. \emph{Top-right}: Environment with obstacles displaced. \emph{Bottom}: Trajectory executions on a real robot in the obstacle-free environment.}}
  \vspace{-0.5cm}
  \label{fig:results_reaching}
\end{figure}

Fig.~\ref{fig:results_reaching} shows multiple instances of reproduction for the \emph{reaching} skill. The skill is reproduced with \eqref{eq:MAP} by conditioning the learned trajectory prior on the likelihood of starting from a desired initial state and the likelihood of staying clear of arbitrarily placed obstacles. We show the trajectories generated from two different initial states in three different environments. When the obstacles are placed at the same location as the demonstration phase or displaced, the reproduced trajectories from the prior without importance weighting take the longer path to the target around the obstacles. This is because the demonstrations on average took a longer path while avoiding obstacles and the prior shown in Fig.~\ref{fig:prior_reaching}(a) forces the reproduced trajectories to exhibit a similar behavior. For the same reasons, the deviant non-smooth trajectories are also observed when no obstacles are present in the vicinity of the robot in the reproduction environment.

\begin{figure}[!t]
\centering
  \subfloat{%
    \includegraphics[trim={0cm 0cm 0cm 0cm}, clip, width=0.23\textwidth]{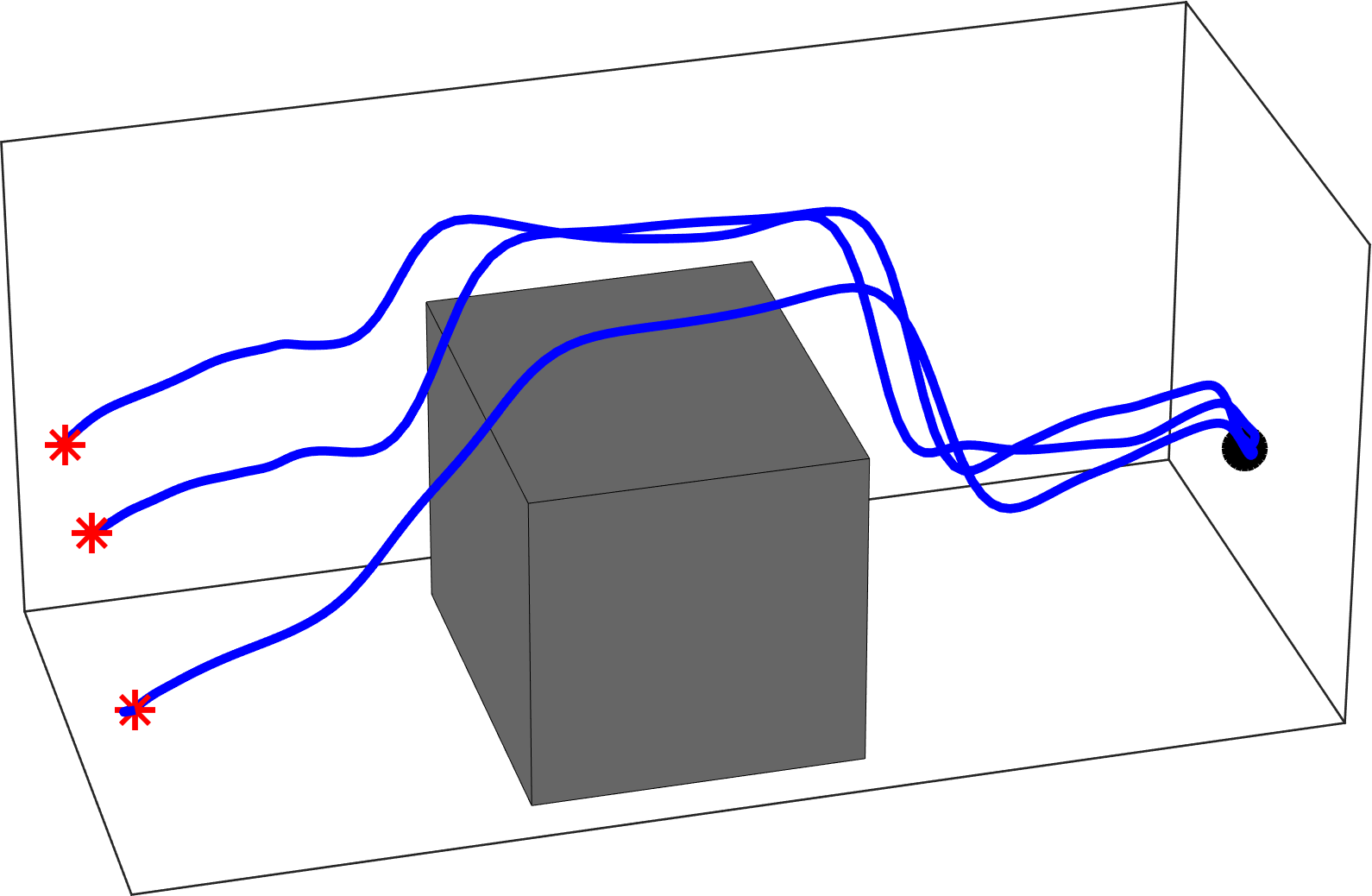}}
    \hfill
  \subfloat{%
    \includegraphics[trim={0cm 0cm 0cm 0cm}, clip, width=0.23\textwidth]{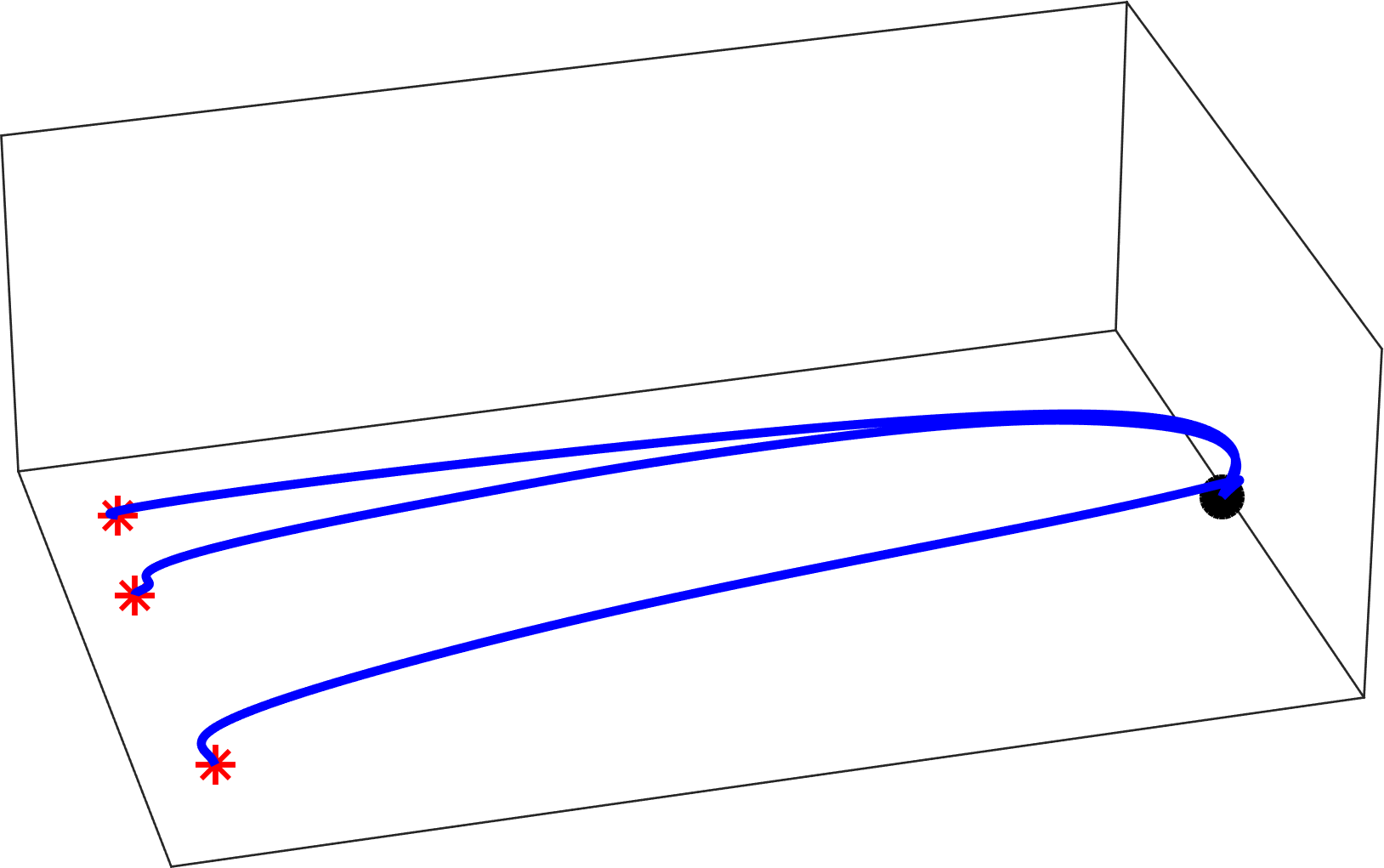}}
	\caption{\small{Human demonstrations for the \emph{placing} skill in two different environments. \emph{Left}: Environment with a large obstacle influencing the demonstrations. \emph{Right}: Obstacle-free environment.}}
	\vspace{-0.6cm}
	\label{fig:demos_placing}
\end{figure}

The \textit{placing} skill involves placing an object at different locations on a table. All the demonstrations start from the same location since the object's initial location is fixed. The end state of the demonstration varies with the target placement location. Initially there is an obstacle present in the desired path, hence all the demonstrations go above the obstacle causing them to be influenced. Fig.~\ref{fig:demos_placing} (left) plots the human demonstrations provided in this scenario. Since only the influenced demonstrations are available at this stage, the trajectory prior learned from these demonstrations also encodes the influence of obstacles which is undesirable. However, as the environment changes and more demonstrations are available in a cleaner environment, as shown in Fig.~\ref{fig:demos_placing} (right), the prior is updated using the incremental weighted learning procedure described in Section \ref{sec:incremental_learning}. 

\begin{figure}[!t]
\centering
  \subfloat{%
    \includegraphics[trim={0cm 0cm 0cm 0cm}, clip, width=0.23\textwidth]{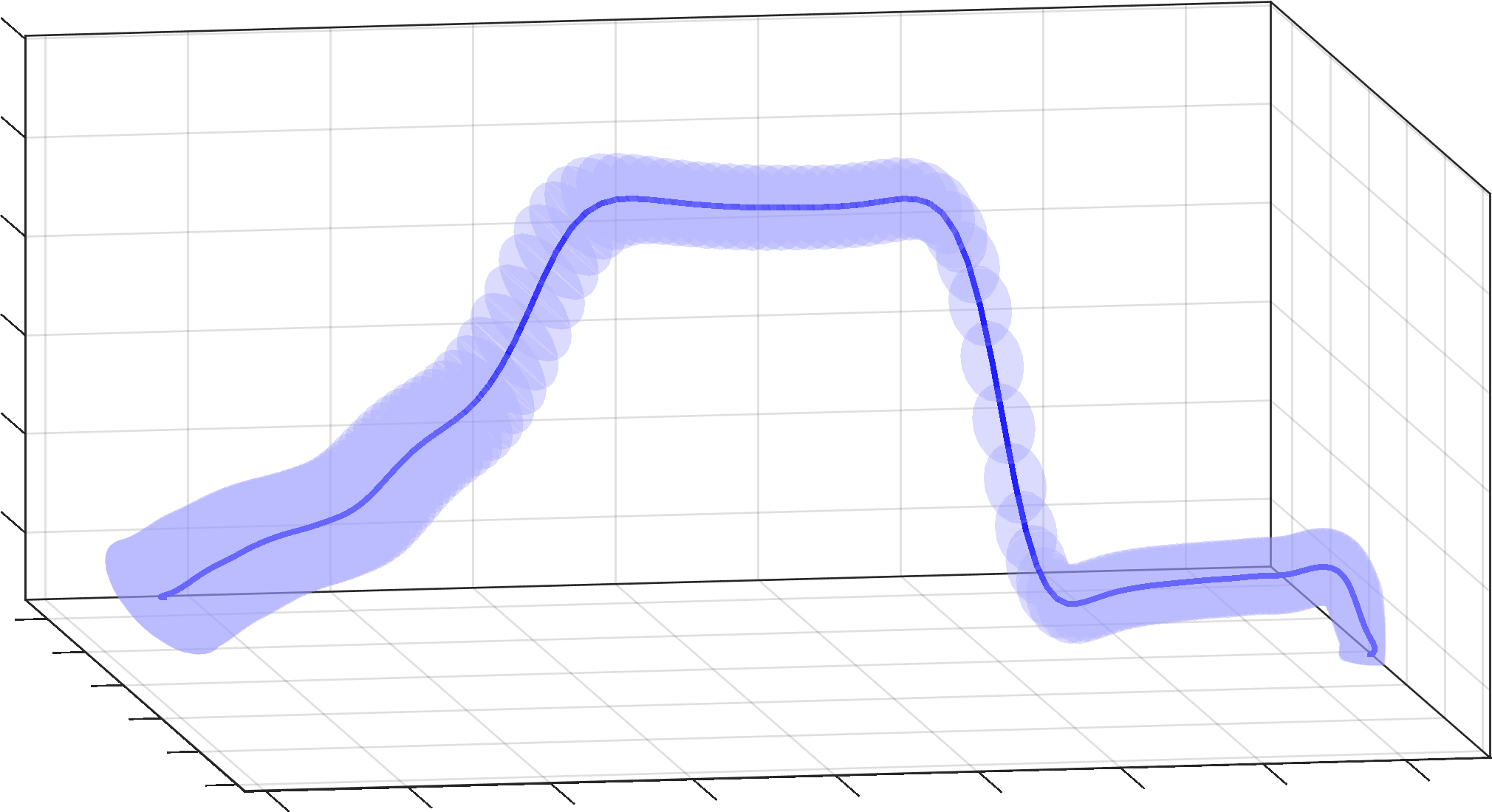}}
    \hfill
  \subfloat{%
    \includegraphics[trim={0cm 0cm 0cm 0cm}, clip, width=0.23\textwidth]{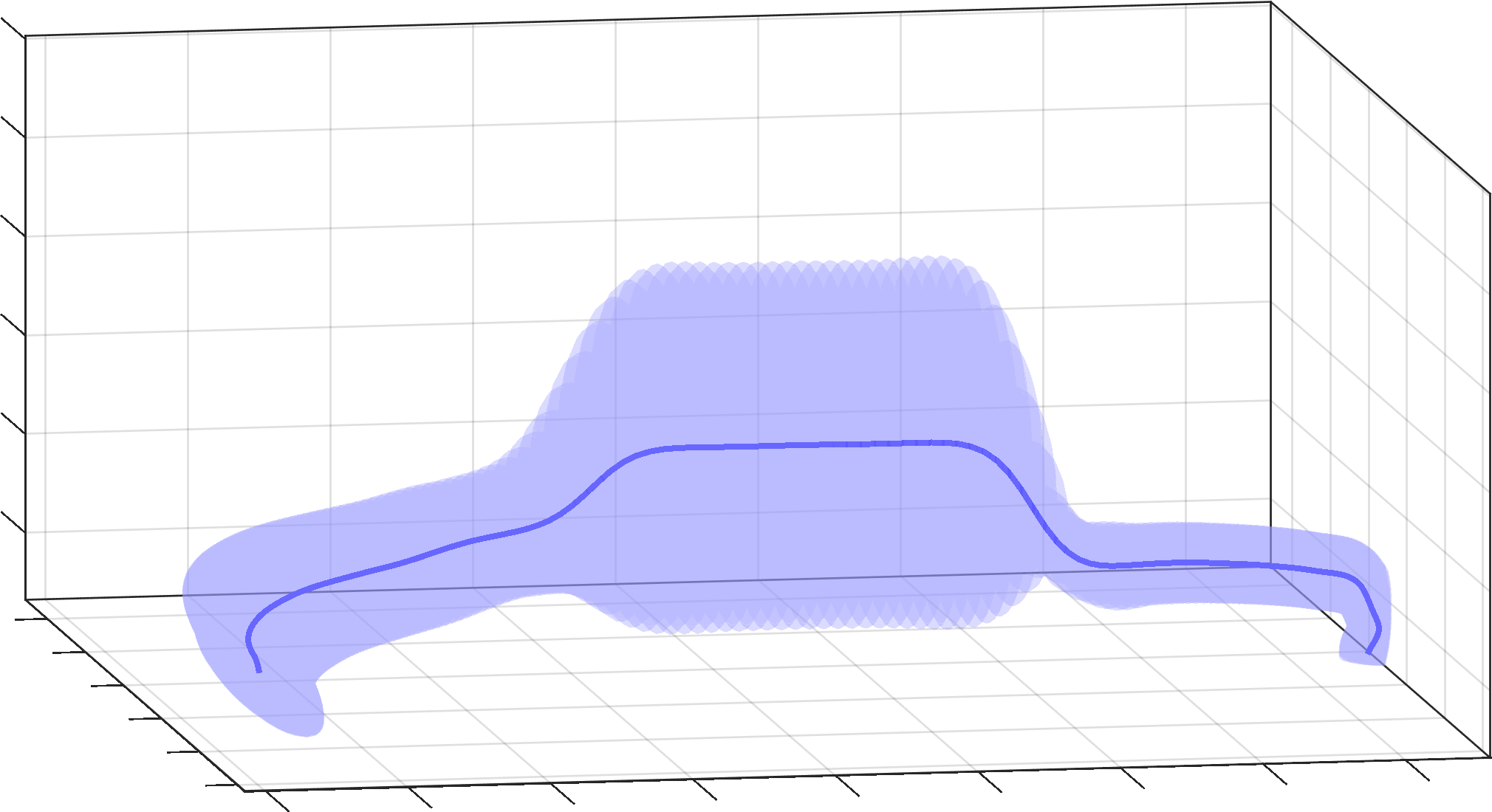}}\\
  \subfloat{%
    \includegraphics[trim={0cm 0cm 0cm 0cm}, clip, width=0.23\textwidth]{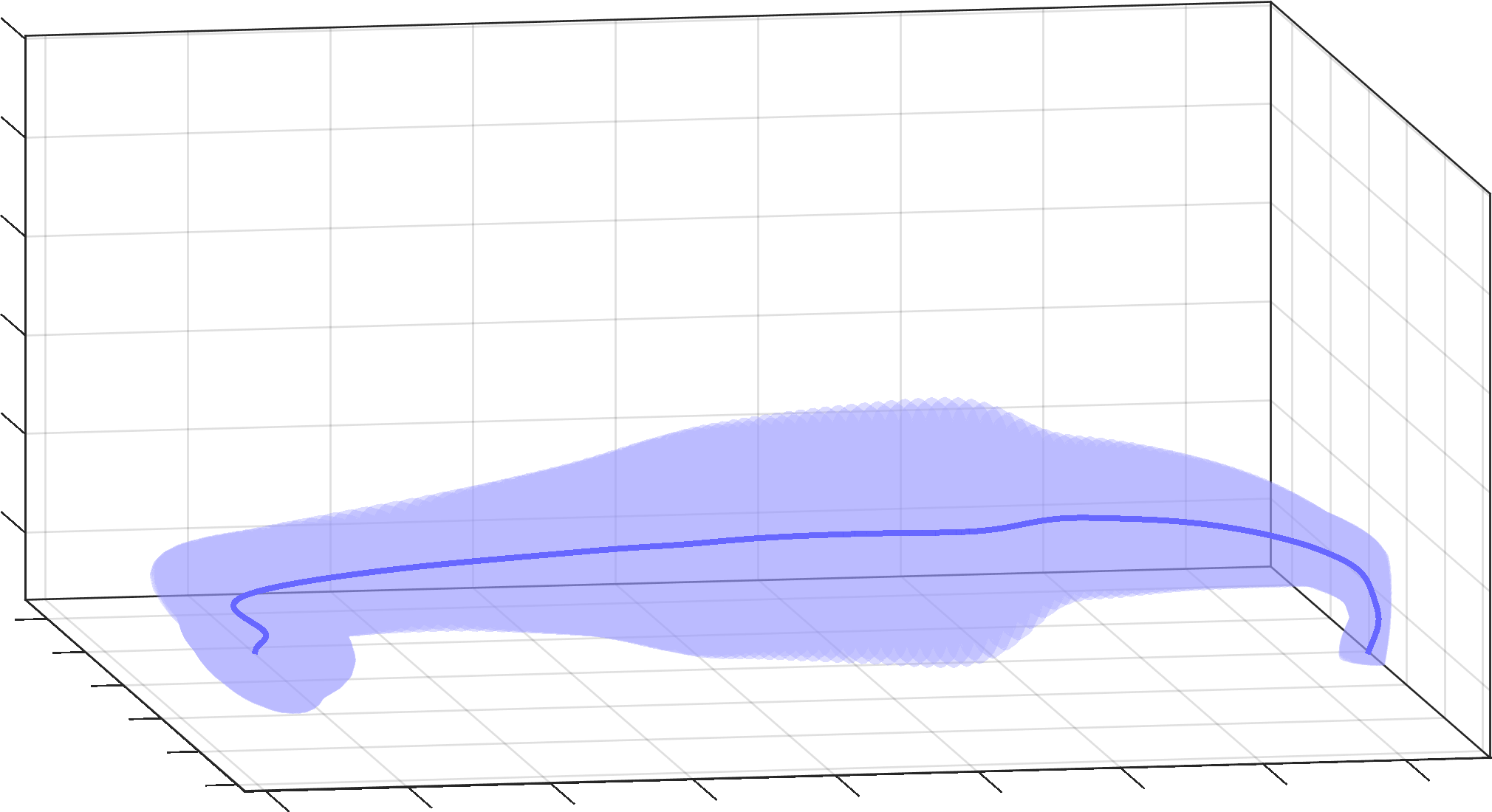}}
    \hfill
  \subfloat{%
    \includegraphics[trim={0cm 0cm 0cm 0cm}, clip, width=0.23\textwidth]{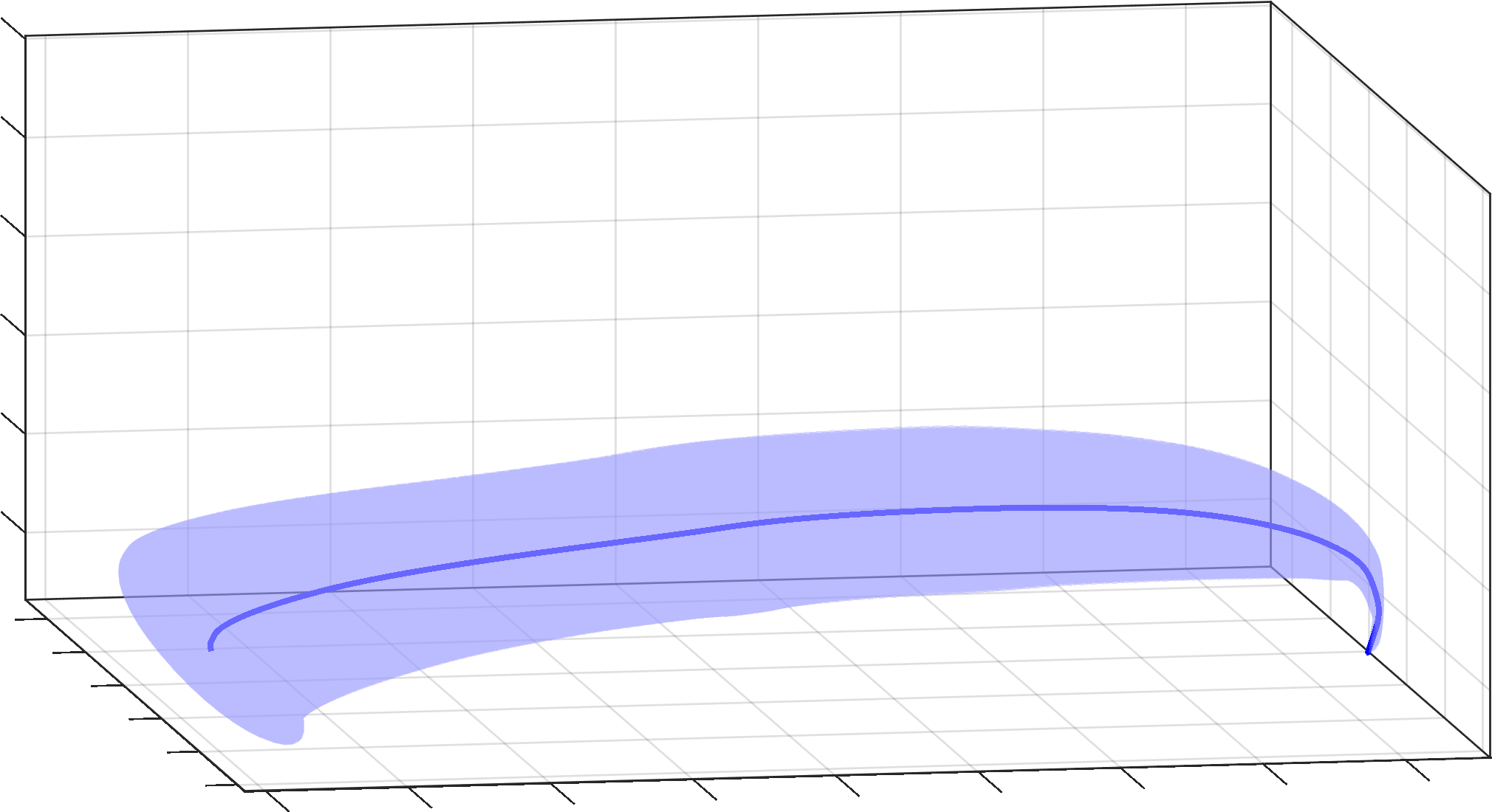}}
	\caption{\small{Trajectory priors for the \emph{placing} skill with importance weighting. \emph{Top-left}: Learned from first 3 demonstrations recorded in the presence of obstacle. \emph{Top-right}: Prior after assimilating fourth demonstration in clean environment. \emph{Bottom-left}: Prior after assimilating fifth demonstration. \emph{Bottom-right}: Final prior after all the incremental updates.}}
	\vspace{-.6cm}
	\label{fig:prior_weighted_placing}
\end{figure}

\begin{figure}[!t]
\centering
  \subfloat{%
    \includegraphics[trim={0cm 0cm 0cm 0cm}, clip, width=0.23\textwidth]{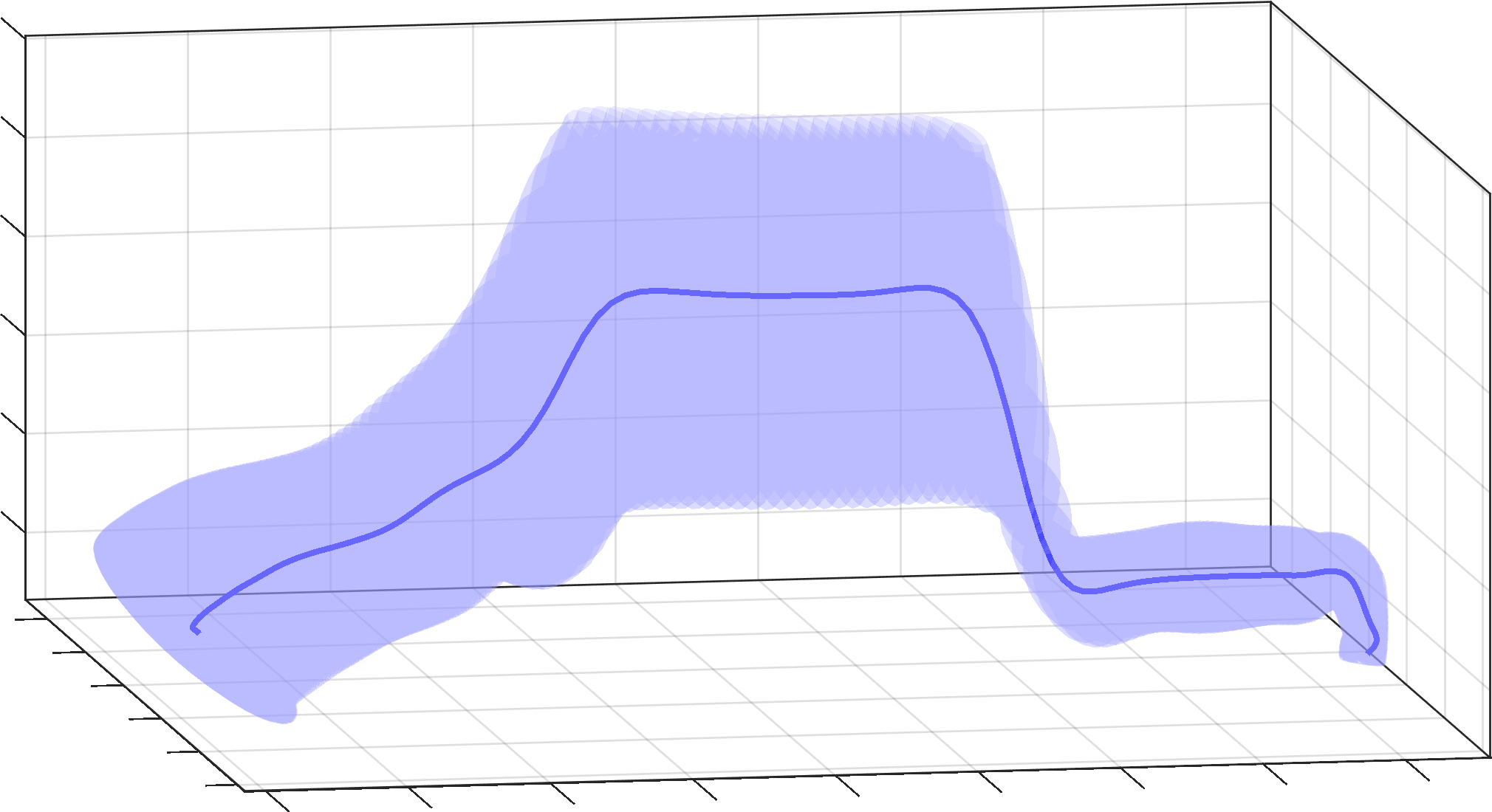}}
	\hfill
  \subfloat{%
    \includegraphics[trim={0cm 0cm 0cm 0cm}, clip, width=0.23\textwidth]{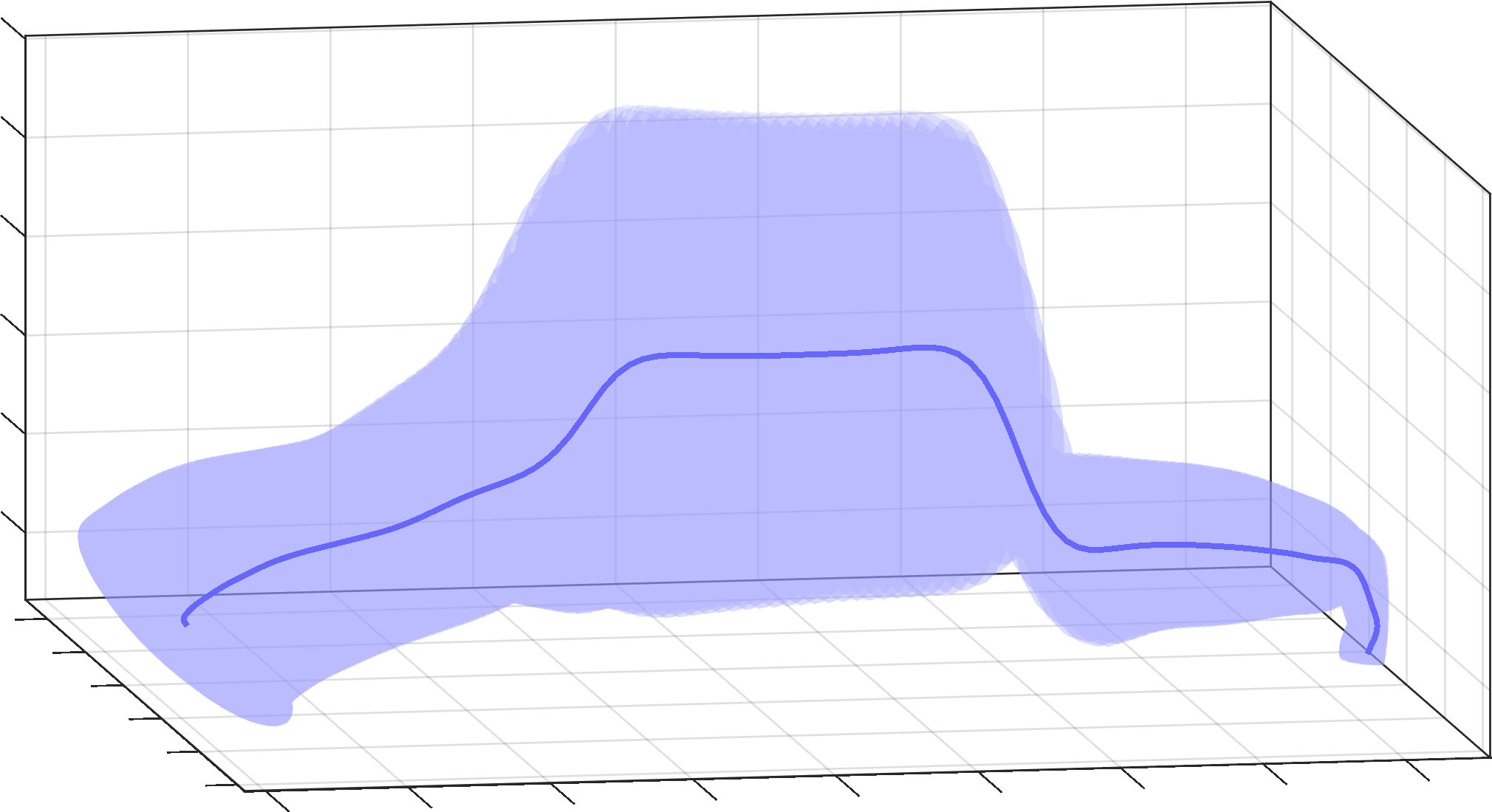}}
     \\
	\caption{\small{Trajectory priors for the \emph{placing} skill without importance weighting. \emph{Left}: Learned after assimilating first 4 demonstrations. \emph{Right}: Final prior after all the incremental updates.}}
	\vspace{-0.6cm}
	\label{fig:prior_unweighted_placing}
\end{figure}

Fig.~\ref{fig:prior_weighted_placing} shows the evolution of the prior as demonstrations are assimilated. The prior initially enforces highly constrained motion causing the trajectories to avoid the obstacle even when it is not present. As more demonstrations are made available in an obstacle-free environment, the high importance weight  relative to the influenced demonstrations enables adaptation to the desired underlying motion after just three updates. On the other hand, when the importance weighting is not considered in the incremental learning procedure, the trajectory prior still exhibits the obstacle influence even after all the demonstrations are incorporated. This is shown in Fig.~\ref{fig:prior_unweighted_placing}. The utility of the incremental learning procedure is high in such scenarios. It is undesirable to keep all the demonstrations and re-learn the prior on arrival of each new demonstration, since this can be both time-consuming as well as memory-intensive.


\section{Conclusion}
We have presented \emph{importance weighted skill learning}, which is a novel technique for learning skills from demonstrations in cluttered environments and generalizing them to new scenarios. Our importance weighting function associates lower weights with parts of demonstrations that are likely to collide with obstacles. We conjecture that demonstrations which are in close proximity to obstacles are more susceptible to not satisfying the constraints of the skill being learned. Hence, those demonstrations should be given lesser importance during the skill learning stage. Our learning approach is also capable of incrementally updating and refining the skill model to incorporate new demonstrations without the need to relearn the model from scratch. Since our learning method is based on extracting the underlying stochastic skill dynamics, it does not share the same disadvantages as approaches that assume a mean trajectory to encode the skill. Furthermore, our reproduction method is capable of generalizing the skill efficiently across various scenarios as demonstrated in the experiments.


\section*{Acknowledgements}
This research is supported in part by NSF NRI 1637758, NSF CAREER 1750483, NSF IIS 1637562, and ONR N00014-16-1-2844. 

\bibliographystyle{IEEEtran}
\bibliography{Rana_IROS_2018}

\end{document}